\documentclass[sigconf]{acmart}
\usepackage{balance}
\usepackage{xcolor}
\newcommand{\MYhref}[3][blue]{\href{#2}{\color{#1}{#3}}}%

\usepackage{subcaption}
\usepackage[ruled,vlined,linesnumbered]{algorithm2e}
\usepackage{flushend}
\def\etal{\emph{et al. }}
\AtBeginDocument{%
  \providecommand\BibTeX{{%
    \normalfont B\kern-0.5em{\scshape i\kern-0.25em b}\kern-0.8em\TeX}}}

\acmSubmissionID{1885}

\begin{document}
\fancyhead{}
\title{Discriminative Latent Semantic Graph for Video Captioning}



\author{Yang Bai$^{1}$, Junyan Wang$^{2}$, Yang Long$^{3}$, {}} 
\author{Bingzhang Hu$^{4}$, Yang Song$^{2}$, Maurice Pagnucco$^{2}$, Yu Guan$^{1}$}
\affiliation{%
  \institution{$^{1}$Open Lab, Newcastle University, Newcastle upon Tyne, UK}
  \institution{$^{2}$School of Computer Science and Engineering, University of New South Wales, Australia}
  \institution{$^{3}$Department of Computer Science, Durham University, Durham, UK}
  \institution{$^{4}$Vision Intelligence Center, Hefei CAS Dihuge Automation Co., LTD}
  \institution{\{y.bai13, yu.guan\}@newcastle.ac.uk,\{junyan.wang, yang.song1\}@unsw.edu.au}
  \institution{hubingzhang@dihuge.com, yang.long@ieee.org, morri@cse.unsw.edu.au}
  \country{}
}


\begin{abstract}
Video captioning aims to automatically generate natural language sentences that can describe the visual contents of a given video. Existing generative models like encoder-decoder frameworks cannot explicitly explore the object-level interactions and frame-level information from complex spatio-temporal data to generate semantic-rich captions.
Our main contribution is to identify three key problems in a joint framework for future video summarization tasks.
\textbf{1) Enhanced Object Proposal}: we propose a novel Conditional Graph that can fuse spatio-temporal information into latent object proposal.  \textbf{2) Visual Knowledge}: Latent Proposal Aggregation is proposed to dynamically extract visual words with higher semantic levels.
\textbf{3) Sentence Validation}: A novel Discriminative Language Validator is proposed to verify generated captions so that key semantic concepts can be effectively preserved.
Our experiments on two public datasets (MVSD and MSR-VTT) manifest significant improvements over state-of-the-art approaches on all metrics, especially for BLEU-4 and CIDEr. Our code is available at \MYhref{https://github.com/baiyang4/D-LSG-Video-Caption}{https://github.com/baiyang4/D-LSG-Video-Caption}.
\end{abstract}



\maketitle


\section{Introduction}
With the tremendous growth of video materials uploaded to various online video platforms, \textit{e.g.} YouTube, research in automatic video captioning has received increasing attention in recent years. Thorough video caption can lead to substantial practical impacts, \emph{e.g.} content-based video retrieval and recommendation.
Despite the remarkable progress of computer vision and natural language processing in video analysis and language understanding, video captioning is still a very challenging task. The task requires exploring not only complex object interactions and relationships at frame-level but also high-level story-line from video sequence. Such a task can be seen as a leap from the recognition to comprehension level.

\begin{figure}[t]
\begin{center}
   \includegraphics[width=0.9\linewidth]{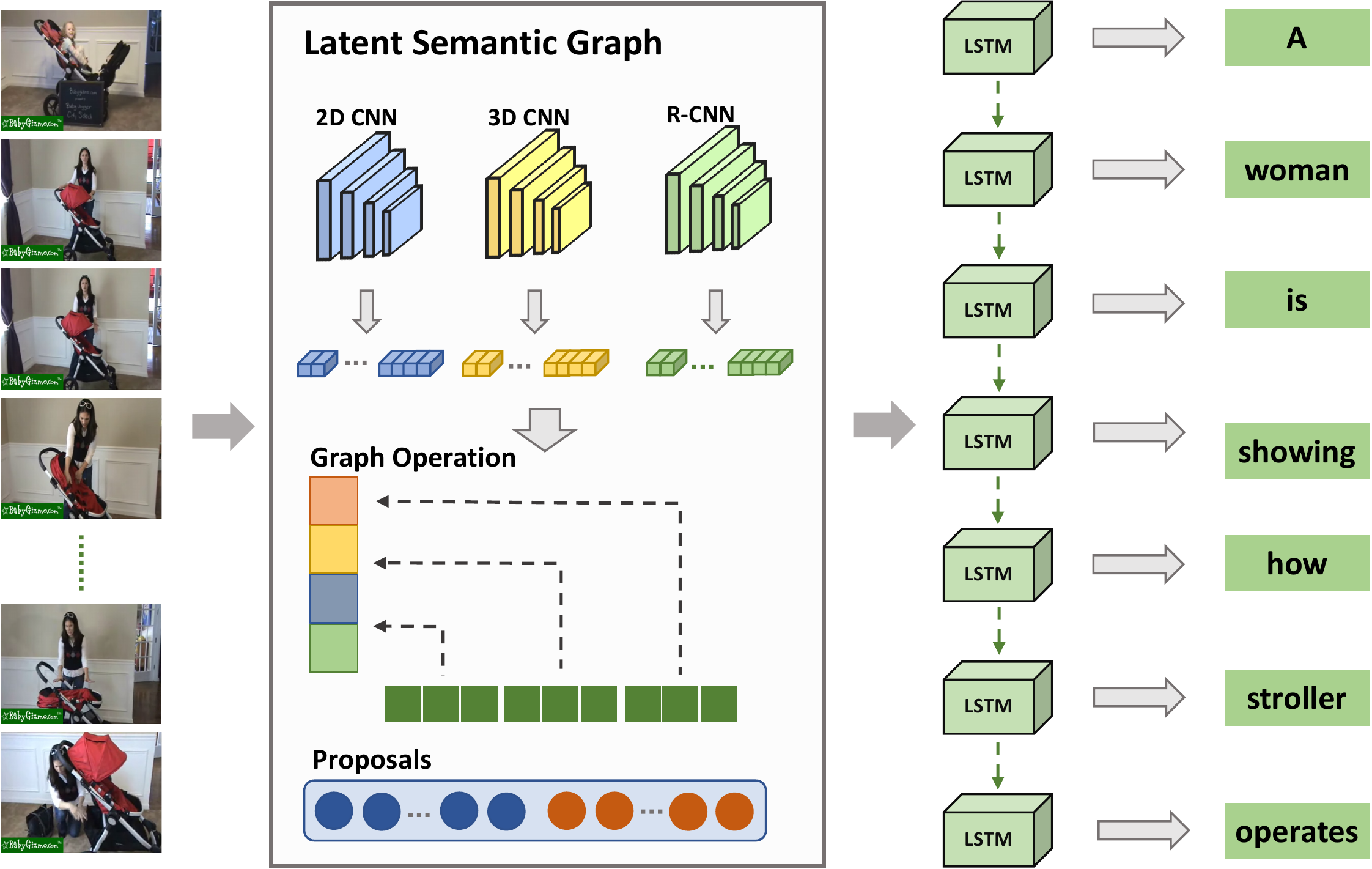}
\end{center}
   \caption{An illustration of the video captioning task. The key challenge is that there is no explicit mapping between video frames and captions. The model needs to jointly consider 2D-CNN, 3D-CNN, and Object proposals from R-CNN and extract high-level semantic visual words to construct a compact caption.}
\label{fig:pipeline}
\end{figure}
One of the main challenges of video captioning is that there is no explicit mapping between video frames and words in captions. The model needs to extract summarized visual words at a much higher semantic level. Figure \ref{fig:pipeline} illustrates an intuitive example of video captioning. From a human perspective, we can interpret the overall process into several sub-tasks:
1) to detect and recognize the main objects in the video, \emph{i.e.,} ``woman'' and ``stroller''; 
2) to infer the action imposed to these objects, \emph{i.e.,} ``showing'' and ``operates'';
3) organizing the contents into a sentence with grammatical structures,  \emph{i.e.,} ``A woman is showing how stroller operates''.
Early studies typically adopt encoder-decoder frameworks ~\cite{venugopalan2015sequence,yao2015describing,wang2018reconstruction} that model video captioning as a machine translation task. These methods focus on modeling the static frame and object features and temporal changes among embedding. To overcome the drawback of embedding-based frameworks, the recent rise of graph neural networks (GNNs) has shown particular advantages in modeling relationships between objects \cite{zhang2020object, yang2018video}. However, applying GNNs to the video captioning task is not a trivial thing. Previous GNNs are mainly built on object features without jointly considering the frame-based spatio-temporal contexts in the entire video sequence. 



The other challenge is that the output caption needs to ensure a grammatical structure keeping content-relevant fidelity rather than producing a list of discrete concepts. To examine whether the expression of a sentence is natural or not, \cite{yang2018video} utilizes a generative adversarial network (GAN) to control the fidelity of generated sentences.
However, video captioning requires a more refined level of supervision to distinguish the real/fake sentences which is conditioned on semantic content, and to ensure grammatical correctness.

The above challenges motivate us to design a new framework for video captioning with three sub-tasks, \emph{i.e.} \textbf{Enhanced Object Proposal, Visual Knowledge,} and \textbf{Sentence Validation}. First, Fusion is about extracting the spatio-temporal contexts from video frames and incorporating such information in the object entities. Note that the number of frames and object proposals in videos is far more than words in captions. Therefore, the second task visual knowledge summary aims to reduce such duplicated and redundant proposals into more compact visual words. 
Such high-level visual words should be easier encoded by a sequential model to produce a caption. The last sentence validation task aims to examine both the fidelity and the readability of the generated caption.
According to the above motivations, we design a \textbf{Discriminative Latent Semantic Graph (D-LSG) }framework with the following insights:
1) \textbf{Graph} model for feature fusion from multiple base models \emph{e.g.} 2D/3D CNN and R-CNN remain unexplored. These features are often heterogeneous in data distribution, dimensions, and structure. 2D CNN represents the frame contents, while 3D CNN extracts the temporal frame changes. We consider such frame-level information as the conditions of all region-level object proposals. Therefore, the conditional graph is not in the traditional form of semi-positive indefinite affinity matrix.
2) \textbf{Latent Semantic} refers to the higher-level semantic knowledge that can be extracted from the enhanced object proposals. Rather than incorporating an external auxiliary knowledge graph as \cite{hou2020joint}, our key idea is to construct a dynamic graph that connects enhanced object proposals with randomly initialized nodes. In other words, the great volume of enhanced object proposals is summarized into high-level visual knowledge via the dynamic graph.
3) \textbf{Discriminative} module is designed as a plug-in language validator.
Generated and ground truth captions can be reconstructed into visual knowledge so as to compare with that extracted from the enhanced object proposals. We adopt the Multimodel Low-rank Bi-linear (MLB) \cite{kim2016hadamard} pooling as metrics to provide finer-level supervision to carry out the sentence validation task.
In summary, our contributions include:
\begin{itemize}
    \item To identify Enhanced Object Proposal, Visual Knowledge, and Sentence Validation sub-tasks in a unified framework for future video summarization tasks.
    \item  A Condition Graph Operation is proposed to enhance region-level object proposal representations with spatio-temporal information of base features of video frames.
    \item Latent Proposal Aggregation with a dynamic graph model is proposed to compress enhanced object proposals into visual knowledge with higher semantic meanings in a latent space.
    \item A Discriminative model is plugged as a validation network that can distinguish generated sentences from ground truth captions and encourage the generated captions to be more content-relevant and semantic-richer.
    \item Quantitative and qualitative experiments on two datasets, MSVD \cite{chen2011collecting} and MSR-VTT \cite{xu2016msr}, demonstrate significant performance boost on all evaluations while achieving significant performance improvement on CIDEr.
    
\end{itemize}

\section{Related Work}
\noindent\textbf{Video Captioning.}
As a joint field in Computer Vision (CV) and Natural Language Processing (NLP), video captioning has received increasing research interest.
Early research on video captioning mainly focused on template-based language models~\cite{guadarrama2013youtube2text,rohrbach2014coherent}. 
With the rapid development of deep learning techniques, recent methods have developed sequence-learning based methods with an encoder-decoder structure \cite{venugopalan2015sequence,yao2015describing,wang2018reconstruction} and considered video captioning as a machine translation task \cite{venugopalan2015sequence}. 
For example, The work of Yao \etal \cite{yao2015describing} adopted a temporal attention mechanism to summarize the visual features for each generated word dynamically.
More recently, object-level information has drawn more attention \cite{zhang2019object,hu2019hierarchical,tan2020learning}.
For instance, the work of \cite{hu2019hierarchical} applied two LSTM layers to construct temporal structures at both frame-level and object-level.
\cite{tan2020learning} proposed a visual reasoning approach on videos over both space and time. 
Zhang \etal \cite{zhang2019object} proposed an object-aware aggregation with bidirectional temporal graph (OA-BTG), which captures detailed temporal dynamics for salient objects in videos.
In summary, these previous methods focus on modeling global information or temporal structure of salient objects. We believe our proposed D-LSG framework initialised three new research questions for video captioning. First, manipulation of interaction between different objects via graph modeling will lead to a new research trend in video captioning. Our proposed conditional graph operation is so far the first approach that can incorporate heterogeneous features from multiple base models. Second, how to extract visual knowledge from enhanced object proposals shares the same spirit of the traditional Bag-of-Visual-Words (BoVW) paradigm. The dynamic graph can be seamlessly included in end-to-end training and is much more cost-efficient. Finally, we point out that the sentence validation requires finer supervision in the discriminative model so that the fidelity and sentence structure can be both preserved.  

\begin{figure*}
\begin{center}
\includegraphics[width=0.9\linewidth]{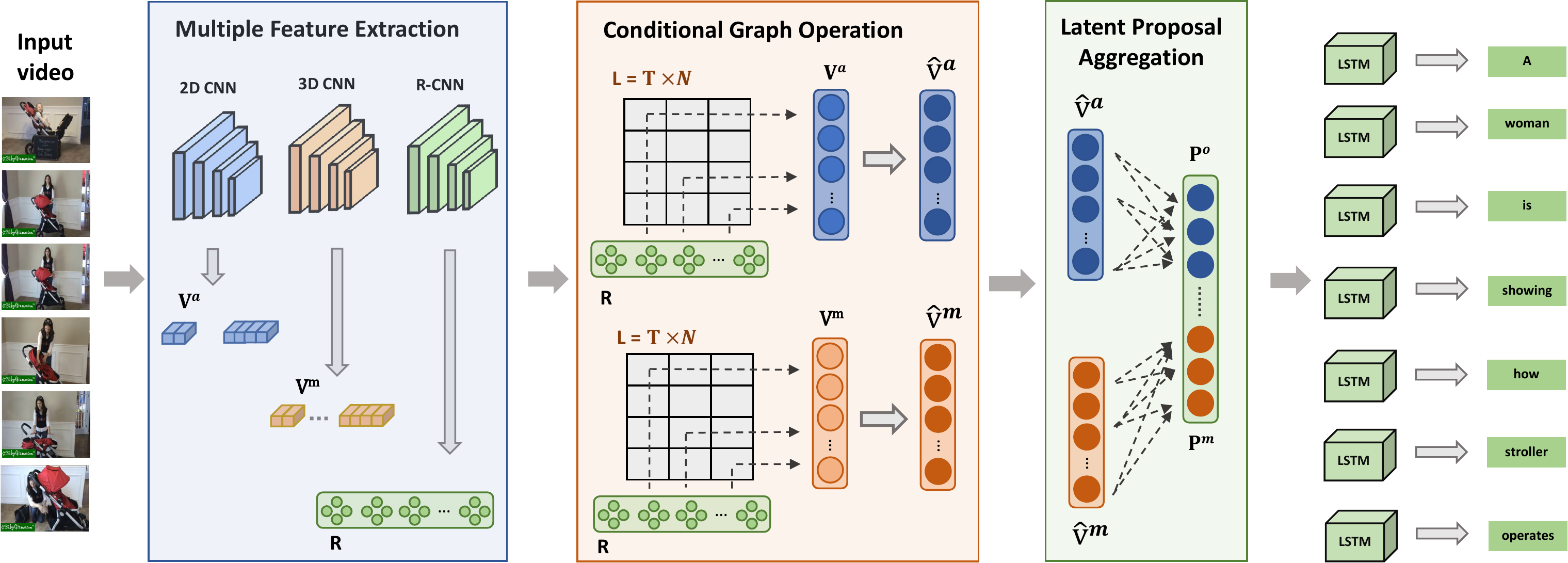}
\end{center}
   \caption{Overview of proposed LSG framework. Base features from 2D/3D CNN and R-CNN provide object and contexts features at frame and region levels. Conditional Graph Operation is applied to appearance and motion channels to compute Enhanced Object Proposals $\hat{V}^a$ and $\hat{V}^m$. $T$ frames of $\hat{V}^a$ and $\hat{V}^m$ are selected into $K$ Visual Knowledge before LSTM captioning.}
\label{fig:overview}
\end{figure*}

\noindent\textbf{Discriminative modeling.}
So far, many studies have investigated image and video captioning using discriminative modeling. 
The work of \cite{shetty2017speaking} applied a discriminator model to distinguish ground truth captions from generated captions, which relies on the Gumbel-Softmax approximation \cite{jang2016categorical}, whereas \cite{dai2017towards} utilizes the policy gradient and their discriminator focuses on caption naturalness and image relevance.
The work of \cite{park2019adversarial} designed a multi-discriminator system that encourages better multi-sentence video description.
However, discriminative modeling for caption generation suffers from stability issues and requires pre-trained generators.
Thus, instead of using the Gumbel-Softmax or policy gradient based method, we propose a semantic relevance discriminative graph based on Wasserstein gradient penalty \cite{gulrajani2017improved}, which can directly feed output and graph-based latent semantic concepts from the generator and do not need to pre-train the generator. The proposed discriminative encoder can be plug into an end-to-end model to reconstruct captions into visual knowledge so that the fidelity and sentence structure can be validated and key semantic entities can be preserved.

\noindent\textbf{Graph Neural Networks.}
Efficient and practical neural network algorithms for processing graph-structured data have become one of the most important machine learning subareas.
Recently, pioneer work~\cite{zhang2020object,pan2020spatio,hou2020joint} have tried to adopt graph neural networks to video captioning. 
For example, Pan \etal \cite{pan2020spatio} proposed a spatio-temporal graph model for video captioning that exploits object interactions in space and time.
Another graph based video captioning research \cite{zhang2020object} proposed an object relational graph based encoder, which captures more detailed object interaction features to enrich visual representation. However, the weights they used to summarize object features in temporal space are the same as the frame features, which may lead to degraded temporal information in summarized object features. In comparison, our conditional graph jointly considers object, contexts, and motion information at both region and frame levels.
The work of \cite{hou2020joint} proposed a joint commonsense and relation reasoning method that applies auxiliary databases for pre-training knowledge graphs as prior knowledge for image and video captioning.
Our dynamic graph in the Latent Proposal Aggregation module is able to extract high-level latent semantic concepts without an external dataset for training.

\section{Methodology}
The video captioning problem is essentially modeled as a sequence to sequence process.
Formally, given a sequence of $T$ frames from video $\boldsymbol{X} = \{\boldsymbol{x}_1,\dots, \boldsymbol{x}_T\}$,
we aim to build an end-to-end model to generate the caption $\boldsymbol{Y} = \{\boldsymbol{y}_1, \dots, \boldsymbol{y}_{T'}\}$ for the given video. Note $T\neq T'$, which forms an open task that is even very challenging for humans. In this paper, we identify three key sub-tasks, namely Enhanced Object Proposal, Visual Knowledge, and Sentence Validation, details of which are introduced as follows.

\subsection{Architecture Design}
\begin{figure}[t]
\begin{center}
    \includegraphics[width=0.9\linewidth]{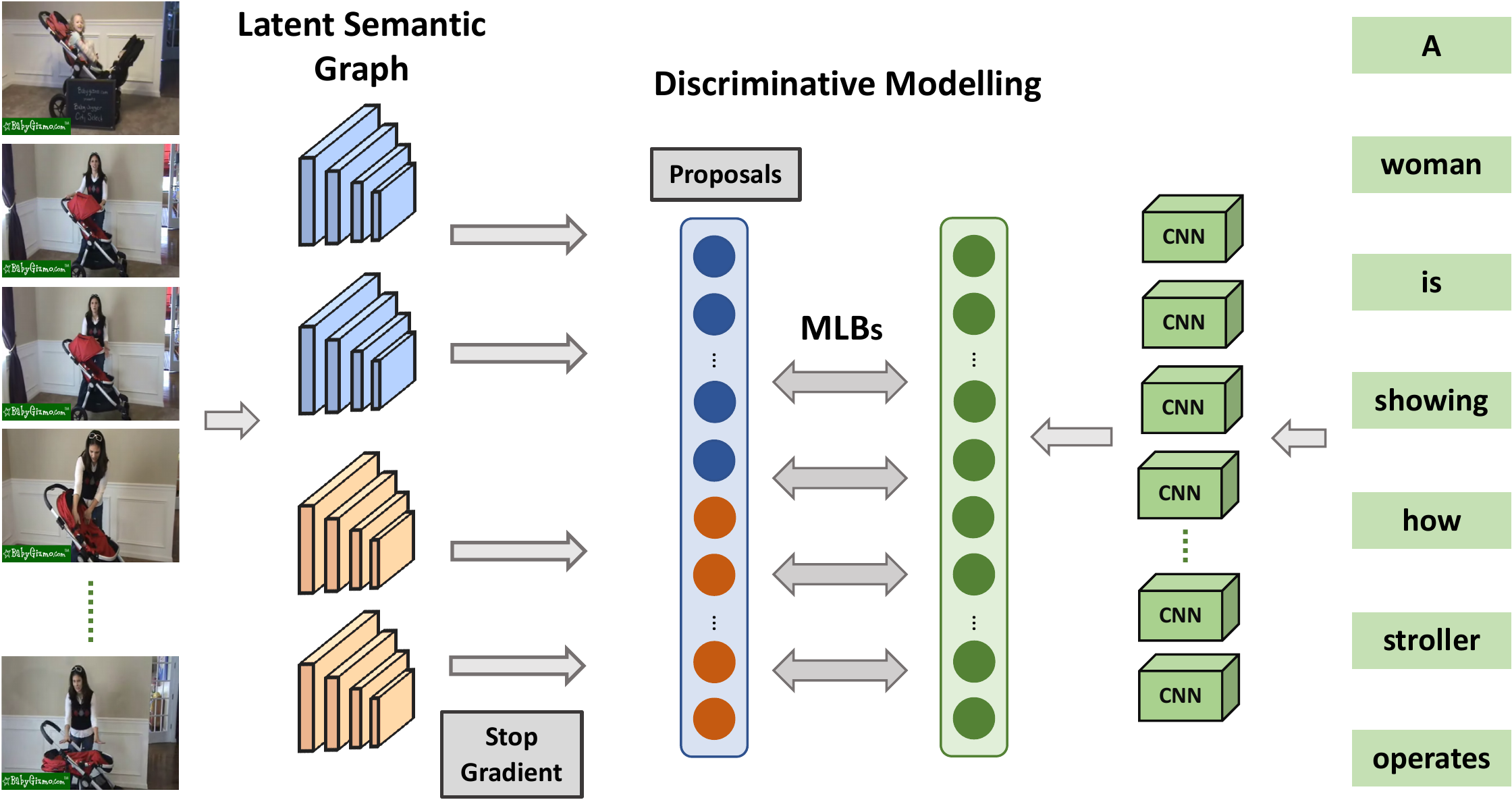}
\end{center}
   \caption{An overview of our discriminative modeling process. We score an input sentence in a semantic concept perspective of view. The model reconstructs the visual knowledge based on the input sentence and output comparison scores with visual knowledge encoded from the corresponding video.}
\label{fig:long}
\end{figure}

The overview of our proposed model is illustrated in Figure \ref{fig:overview}. 
The Latent Semantic Graph (LSG) consists of three parts: (1) Multiple Feature Extraction; (2) Conditional Graph Operation; (3) Latent Proposal Aggregation. A plug-in discriminative caption validator is illustrated in Figure \ref{fig:long}. We introduce how our proposed models can address the sub-tasks next.

\noindent\textbf{Multiple Feature Extraction}.
Given input video frames $\boldsymbol{X}$, the model first extracts visual context representations. In this work, 2D CNNs and 3D CNNs are employed to extract \emph{appearance features} $\boldsymbol{V}^{a} = \{\boldsymbol{v}^{a}_t\}^T_{t=1}$ and \emph{motion features} $\boldsymbol{V}^{m} = \{\boldsymbol{v}^{m}_t\}^T_{t=1}$ respectively.
Object proposals are extracted by R-CNNs to capture key entities with \emph{region features} $\boldsymbol{R} = \{\boldsymbol{r}_{t}\}^T_{t=1}$ from each frame, where
$\boldsymbol{r}_{t} = \{\boldsymbol{r}_{t}^i \in \mathbb{R}^{D_r}\}^N_{i=1}$ and $N$ denotes the number of region features in each frame. 
Thus, the total number of object proposals is denoted as $L = T \times N$.
Next, to make full use of motion information, we concatenate the appearance features and motion features, and apply LSTM models to learn better representations of motion features.

\noindent\textbf{Enhanced Object Proposal}. In video captioning, one of the essential tasks is to detect and recognize the entities.
The weak object proposals in region feature $\boldsymbol{R} \in \mathbb{R}^{T \times N \times D_r}$ are enhanced by their visual contexts of appearance and motion, respectively, which result in \emph{enhanced appearance proposals} $\boldsymbol{\hat{V}}^a \in \mathbb{R}^{T \times D_g}$ and \emph{enhanced motion proposals} $\boldsymbol{\hat{V}}^m \in \mathbb{R}^{T \times D_g}$ in a graph structure, where $D_g$ is the feature dimension used in graph operation. $\boldsymbol{\hat{V}}^a$ and $\boldsymbol{\hat{V}}^m$ together form the enhanced object proposals.

\noindent\textbf{Visual Knowledge}.
The Latent Proposal Aggregation (LPA) module introduces a dynamic graph that can summarize the enhanced appearance and motion features to latent semantic proposals as $K$ dynamic \emph{visual words}:
$\boldsymbol{P}^{o} \in \mathbb{R}^{K \times D_g}$ and 
$\boldsymbol{P}^{m} \in \mathbb{R}^{K \times D_g}$. Note $K\ll T$.

\noindent\textbf{Language Decoder}.
Visual knowledge extracted by the LPA is then used to generate corresponding captions.
We adopt the language generation decoder that are commonly used in VQA and  video captioning fields ~\cite{anderson2018bottom, zhang2019object, hou2020joint, tan2020learning}. The language decoder consists of an \emph{attention LSTM} network for weighting \emph{dynamic visual words} and a \emph{language LSTM} network for caption generation. At each time step, \emph{attention LSTM} takes current word embedding and global visual vector $\bar{\boldsymbol{p}} = [\sum^{K}_{k=1}\boldsymbol{p}^o_k, \sum^{K}_{k=1}\boldsymbol{p}^m_k ] \in \mathbb{R}^{2D_g}$ as input and output current hidden state $\boldsymbol{h}_{t}^{attn}$.The $\boldsymbol{h}_{t}^{attn}$ is then treated as the query of the attention operation to weight sum the object and motion visual words to context feature $\boldsymbol{c}_{t}^{op}, \boldsymbol{c}_{t}^{mp} \in \mathbb{R}^{D_g}$.
The \emph{language LSTM} then takes the current context features and current \emph{attention LSTM} hidden states and output current predicted word probability distribution $\boldsymbol{c} \in \mathbb{R}^{D_{vocab}}$, where $D_{vocab}$ is the vocabulary size. 
\subsection{Latent Semantic Graph}
There has been significant research investigating the dependencies between objects and complex content in generating video captions.
However, learning spatio-temporal dependencies remains a challenging issue.
Compared to conventional spatio-temporal convolution and recursive neural networks, graph models provide a new solution to model dependencies.
In this work, we propose the LSG model that can efficiently encode object-level features from videos as highly summarized \emph{visual words} with higher semantic level. 
To progressively generate the high-level concepts representing visual features, the essential parts of the LSG model are divided into two components: conditional graph operation and latent proposal aggregation.

\noindent\textbf{Conditional Graph Operation}.
In video captioning, one of the key challenges is to model the complex object-level interactions and relationships. Another challenge is to learn informative object-level features that are in context of frame-based background information.
To encode object-level information as highly summarized latent semantic objects and motion \emph{visual words} conditioned on frame-based background information, we first aggregate object-level features into appearance and motion features respectively via graph operation. 

Since we have the object-level \emph{region features}, frame-level \emph{motion features} and \emph{appearance features} after the Multiple Feature Extraction step mentioned in Section 3.1, we build a graph neural network to model object-level interactions, where each region feature $\boldsymbol{r}^j$ out of all $L$ region features is regarded as a node.
To modeling the frame-based conditioning, instead of relying only on the local region features, we take the full picture into account, which takes both frame-level motion and appearance features and object-level \emph{region features}. 
Specifically, we pass messages of the region features to frame-level features at each frame $t$:
\begin{equation}
    \hat{\boldsymbol{v}}_{t}^{a}=\boldsymbol{v}_{t}^{a} + \sum_{j=1}^{L} \mathcal{F}_{kernel}\left(\boldsymbol{v}_{t}^{a}, \boldsymbol{r}^{j}\right) \boldsymbol{W}_a \boldsymbol{r}^{j}~,
\end{equation}
where $\hat{\boldsymbol{v}}_{t}^{a}$ represents the $t^{th}$ representation of \emph{enhanced appearance proposal}, and $\boldsymbol{W}_a \in \mathbb{R}^{D_g \times D_r}$ denotes learnable parameters.
Specially, $\mathcal{F}_{kernel}$ is a kernel function that aims to encode relations between frame-level features $\boldsymbol{v}_{t}^a$ and detailed region features $\boldsymbol{r}^{j}$. 
In this study, we define $\mathcal{F}_{kernel}$ as:
\begin{equation}
    \mathcal{F}_{kernel}\left(\boldsymbol{v}_{t}^{a}, \boldsymbol{r}^{j}\right) = 
    \psi(\boldsymbol{v}_{t}^{a}) \phi(\boldsymbol{r}^{j})^T~,
\end{equation}
where $\psi$ and $ \phi$ are linear functions followed by Tanh activation function. 
This step aims to project features from different modalities to a common feature space and compute the similarity to represent the degree of connectivity between region features and frame-level features in the graph.
Alternatively, the equation can be written as:
\begin{equation}
    \hat{\boldsymbol{V}}^{a} = \boldsymbol{V}^a + \boldsymbol{A}(\boldsymbol{V}^a, \boldsymbol{R} )\boldsymbol{R}\boldsymbol{W}_a^T~,
\end{equation}
where $ \boldsymbol{A} \in \mathbb{R}^{T \times L} = \mathcal{F}_{softmax}(\psi(\boldsymbol{V}^a)  \phi(\boldsymbol{R})^T)$ denotes the relation coefficient matrix between \emph{appearance features} and \emph{region features}.
Meanwhile, region features are aggregated to motion features as $\hat{\boldsymbol{V}}^{m}$ in the same process. 
With $\hat{\boldsymbol{V}}^{a}$ and $\hat{\boldsymbol{V}}^{m}$ that contain object-level information on the condition of frame-level features, we then need to summarize the enhanced proposals to obtain informative semantic concept candidates or proposals with less redundancy.

\begin{figure}[t]
\begin{center}
    \includegraphics[width=0.9\linewidth]{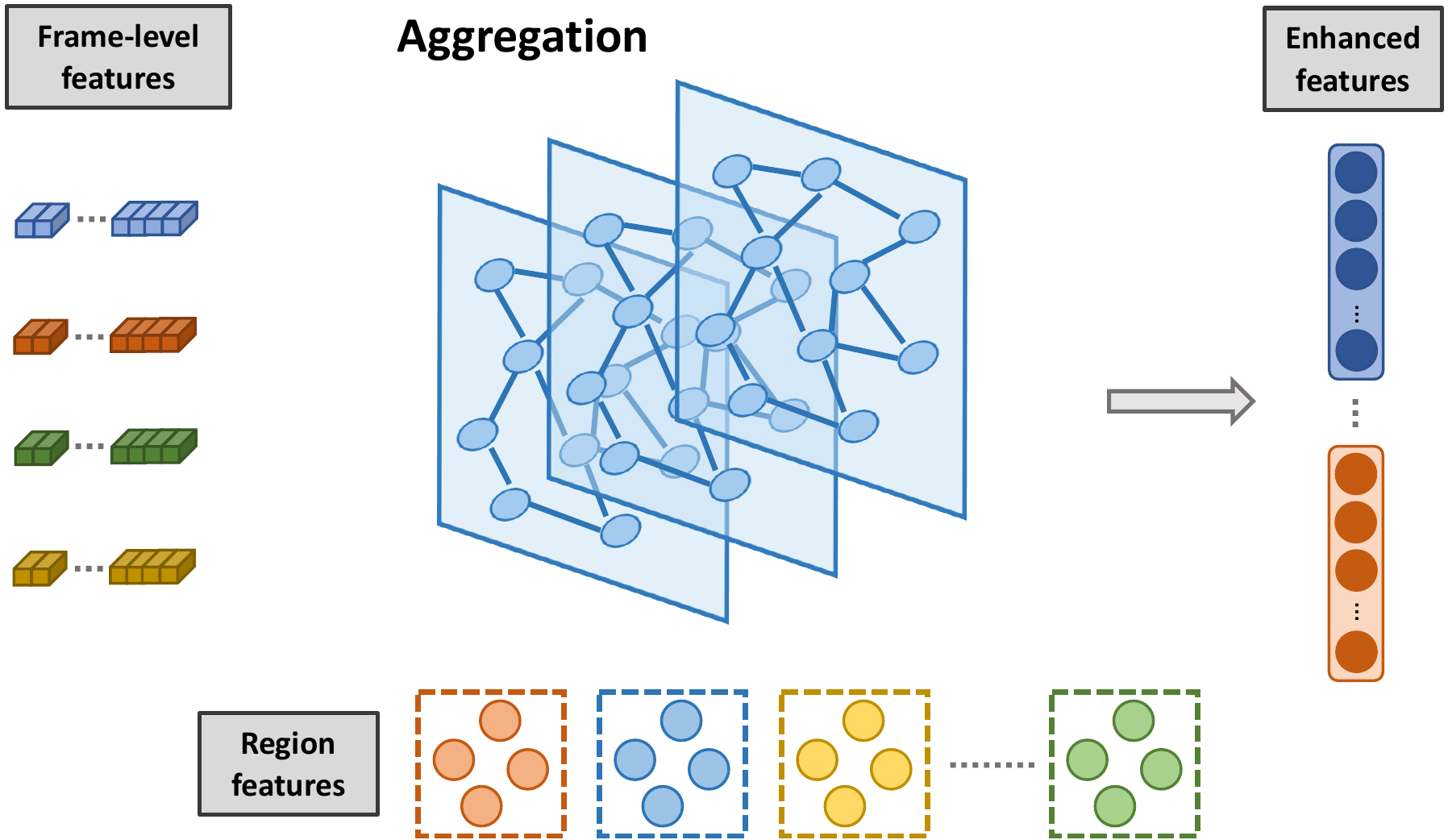}
\end{center}
   \caption{An overview of the aggregation process. 
   }
\label{fig:aggregation}
\end{figure}

\noindent\textbf{Latent Proposal Aggregation}.
To further summarize the \emph{enhanced object proposals}, we propose a latent proposal aggregation method to generate visual words dynamically based on the enhanced features $\hat{\boldsymbol{V}}^{a}$ and $\hat{\boldsymbol{V}}^{m}$ inspired by \cite{zhang2019latentgnn}.
First, we augment the original enhanced proposal nodes $\{\hat{\boldsymbol{v}}^{a}_t\}^T_{t=1}$ and $\{\hat{\boldsymbol{v}}^{m}_t\}^T_{t=1}$ with a set of additional latent nodes, and then aggregate information from the enhanced proposals to the latent nodes in a graph structured manner. 
Specifically, we introduce a set of \emph{object visual words} $\boldsymbol{P}^{o} = \{\boldsymbol{p}^{o}_k \in \mathbb{R}^{D_g} \}^K_{k=1}$, which means potential object candidates in the given video. 
Note that $K$ indicates the number of visual words, so that we can summarize the enhanced proposals into informative dynamic visual words.
The aggregation process is defined as:

\begin{equation}
    \boldsymbol{p}_{k}^{o}=\sum_{j=1}^{T} \mathcal{F}_{kernel}\left(\theta^{o}_{k}, \hat{\boldsymbol{v}}_{j}^{a} \right) \boldsymbol{W}_{op} \hat{\boldsymbol{v}}_{j}^{a}~,
\end{equation}
where $\boldsymbol{p}^{o}_{k}$ denotes the $k^{th}$ object visual word and $\theta^{o}_{k} \in \mathbb{R}^{D_g}$ denotes learnable parameters for the $k^{th}$ object visual word. 
Following the same process, we can derive the \emph{motion visual words} $\boldsymbol{P}^{m} = \{\boldsymbol{p}^{m}_k \in \mathbb{R}^{D_g} \}^K_{k=1}$ that represent potential motion candidates in the given video.
Therefore, with LSG, we extract the high-level representation and summarize information as dynamic visual words from a video that models both object-level interaction and frame-level condition.
The latent semantic visual words are then fed into the language decoder to generate captions as mentioned in Section 3.1. Although the output sequence is generated based on the visual words, there's still potential to obtain video description with more meaningful semantic concepts for more informative caption generation.

\subsection{Discriminative Language Validation}
While other discriminative models for video captioning mainly focus on fluency and visual relevance of the generated descriptions, we aim to generate meaningful captions from the perspective of semantic concepts.
In our approach, we design a discriminative model as a language validation process that encourages the generated captions to contain more informative semantic concepts via reconstructing the visual words or knowledge based on the input sentences under the condition of corresponding true visual words encoded by LSG.
Specifically, based on the visual knowledge $\boldsymbol{P}^o$ and $\boldsymbol{P}^m$ encoded from input video features, we propagate information from the generated captions to reconstruct visual knowledge and discriminate the reconstructed visual words from ground-truth and generated captions in an adversarial training manner.
The process of the discriminative modeling is summarized in Algorithm 1 and described as follows.

Given the output word sequence from the language decoder, the discriminative model aims to distinguish the generated captions and ground-truth with regard to the semantic concepts in the corresponding sentences. 
To prevent the discriminative model from easily distinguishing between real and fake samples without learning useful information (the ground truth caption is the one-hot integer datatype while the generated caption contains probability distributions) and to stabilize the training process, we employ the WGAN-GP architecture as it uses the earth-mover distance to capture the difference between real and fake samples, which is well suited to our problem.

The discriminative model first extracts sentence features $\boldsymbol{S}  \in \mathbb{R}^{T'  \times D_s}$ from the given caption $\boldsymbol{C} \in \mathbb{R}^{T'\times D_{vcab}}$ by using several layers of 1D CNN with residual connections.
Then, we adopt a graph-based structure to obtain the reconstructed object and motion visual words $\tilde{\boldsymbol{P}}^o$ and $\tilde{\boldsymbol{P}}^m$ from the caption:
\begin{equation}
    \tilde{\boldsymbol{P}}^o = \boldsymbol{A}(\boldsymbol{P}^o, \boldsymbol{S} )\boldsymbol{S}\boldsymbol{W}_{oc}~.
    \label{eq:ag latent proposal}
\end{equation}
Note that visual words $\boldsymbol{P}$ from the LSG model stops gradient descent before passing to the discriminative model so that it does not affect the caption generation.
Following the same process, we can derive the aggregated motion visual words $\tilde{\boldsymbol{P}}^m$.
Then, we compare the visual words $\boldsymbol{P}$ with the aggregated visual words $\tilde{\boldsymbol{P}}$ by Multimodal Low-rank Bi-linear pooling (MLB)
\cite{kim2016hadamard}, which is recognized to be efficient in tasks such as VQA.
To be specific, for each pair of visual words $\tilde{\boldsymbol{p}}_i$ and $\boldsymbol{p}_i$:
\begin{equation}
    d_{i}=\sigma\left(\tanh \left(\boldsymbol{U}^{T} \tilde{\boldsymbol{p}}_{i}\right) \odot \tanh \left(\boldsymbol{V}^{T} \boldsymbol{p}_{i}\right)\right)~,
\end{equation}
where $\sigma$ is the sigmoid activation function, $\odot$ denotes the Hadamard product, $\boldsymbol{U}$ and $\boldsymbol{V}$ are learnable parameters. Note that $d_{i}$ is a scalar which means the score of the input aggregated proposal compared to the original one. $d_{i}^o$ is score of $i^{th}$ object visual word pair and $d_{i}^m$ is score of $i^{th}$ motion visual word pair.
Then the overall comparison score can be expressed as:
\begin{equation}
    d^{o} = \frac{1}{K'}\sum_{i=1}^{K'}d^o_i~, d^{m} = \frac{1}{K'}\sum_{i=1}^{K'}d^m_i~,
\end{equation}
where $K'$ is the number of visual words selected out of $K$ for reconstruction and comparison.
The intuition behind is that the captions contain less semantic information than video.
Instead of adding $d^{o}$ and $d^{m}$ as the discriminative model's output, we weight them adaptively based on the sentence feature since sentences have different proportions of object and motion concepts. We calculate the output of the discriminative model as following:
\begin{equation}
    \beta_s = \frac{e^{a^T_o \bar{S}}}{e^{a^T_o \bar{S}} + e^{a^T_m \bar{S}} }~,
\end{equation}
\begin{equation}
    D(\boldsymbol{C}|\boldsymbol{P}) = \beta_s d^{o} + (1-\beta_s) d^{m}~,
\end{equation}
where $a_o, a_m  \in \mathbb{R}^{D_s}$ are learned parameters, and $\bar{S} \in \mathbb{R}^{D_s}$ is mean pooled sentence feature.
$D(\boldsymbol{C}|\boldsymbol{P})$ is the output of the discriminative model that learns to give real captions large values and minimize the values of generated captions. 
For the ground-truth caption $\boldsymbol{C}^r$ and generated caption $\boldsymbol{C}^g$, the loss function of the discriminative model is defined as:
\begin{equation}
    \mathcal{L}_D = D(\boldsymbol{C}^g | \boldsymbol{P}) - D(\boldsymbol{C}^r | \boldsymbol{P}) + \lambda (\left\|\nabla_{\hat{\boldsymbol{C}}} D(\hat{\boldsymbol{C}})\right\|_{2}-1)^{2}~,
\end{equation}
where $\hat{\boldsymbol{C}}$ is sampled along straight lines between real caption $\boldsymbol{C}^r$ and generated caption $\boldsymbol{C}^g$. 
$(\left\|\nabla_{\hat{\boldsymbol{C}}} D(\hat{\boldsymbol{C}})\right\|_{2}-1)^{2}$ is the gradient penalty term that forces the gradient from the generated sample to the real sample to be as small as possible to satisfy the Lipschitz constraint \cite{gulrajani2017improved}.
Thus, for the generator, the loss function is calculated as:
\begin{equation}
\begin{split}
    \hat{\mathcal{L}_G} = -D(\boldsymbol{C}^g | \boldsymbol{P})~,\\
    \mathcal{L}_G = \mathcal{L}_C + \beta \hat{\mathcal{L}_G}~,
    \end{split}
\end{equation}
where $\beta$ is the hyperparameter that controls the weight of $\hat{\mathcal{L}_G}$. $\mathcal{L}_C$ is the caption generation loss.

\begin{algorithm}
\DontPrintSemicolon
\SetKwInOut{Require}{Require}
\SetKwFunction{FDisc}{$\mathcal{F}_{Disc}$}
 \SetKwInOut{Initialize}{Initialize}
 \SetKwProg{Fn}{Function}{:}{}
 \tcc{$\theta^{Disc}$ : Parameters of Discriminative model;}
 \tcc{$n_{Disc}$ :  Number of Discriminative model iterations per generator iteration;}
 \tcc{$\theta^{LSG}$ : Parameters of LSG model;}
 \tcc{$\mathcal{F}_{SG}$ : Stop Gradient;}
 
 \Fn{\FDisc{$\boldsymbol{P}^o$, $\boldsymbol{P}^m$, $\boldsymbol{C}$}}{
        \Require{object visual words $\boldsymbol{P}^o$; motion visual words $\boldsymbol{P}^m$; word sequence $\boldsymbol{C}$.}
        $\boldsymbol{S} = CNNs(\boldsymbol{C} )$ \;
        $\boldsymbol{P}^o$, $\boldsymbol{P}^m$ = $\mathcal{F}_{SG}(\boldsymbol{P}^o)$,  $\mathcal{F}_{SG}(\boldsymbol{P}^m)$ \;
        $\tilde{\boldsymbol{P}}^o = \boldsymbol{A}(\boldsymbol{P}^o, \boldsymbol{S} )\boldsymbol{S}\boldsymbol{W}_{oc} $, $\tilde{\boldsymbol{P}}^m = \boldsymbol{A}(\boldsymbol{P}^m, \boldsymbol{S} )\boldsymbol{S}\boldsymbol{W}_{mc} $\;
        $d_{k}=\sigma\left(\tanh \left(U^{T} \hat{\boldsymbol{p}}_{k}\right) \odot \tanh \left(V^{T} \boldsymbol{p}_{k}\right)\right)$ \;
        $d^{o} = \frac{1}{K'}\sum_{k=1}^{K'}d^o_k~$ \;
        $d^{m} = \frac{1}{K'}\sum_{k=1}^{K'}d^m_k$ \;
        $\beta_s = \frac{e^{a^T_o S}}{e^{a^T_o S} + e^{a^T_m S} }$ \;
        \KwRet $\beta_s d^{o} + (1-\beta_s) d^{m}$\;
  }

 \Initialize{$\theta^{LSG}$, $\theta^{Disc}$}
 \For{i = 1 to epoch number}{
   Sample $i^{th}$ minibatch of video $\boldsymbol{V}_i$ and corresponding caption $\boldsymbol{C}_i^r$ \;
   $\boldsymbol{P}^o_i$, $\boldsymbol{P}^m_i$, $\boldsymbol{C}^g_i$ = $\mathcal{F}_{LSG}(\boldsymbol{V}_i)$ \;
   \For{t = 1 to $n_{Disc}$}    
        { 
        	$ \mathcal{L}_D = \mathcal{F}_{Dis}(\boldsymbol{P}^o_{i}, \boldsymbol{P}^m_{i}, \mathcal{F}_{SG}(\boldsymbol{C}^g_{i})) -  \mathcal{F}_{Dis}(\boldsymbol{P}^o_{i}, \boldsymbol{P}^m_{i}, \boldsymbol{C}_i^r) + \lambda (\left\|\nabla_{\hat{C}}  \mathcal{F}_{Dis}(\boldsymbol{P}^o_{i}, \boldsymbol{P}^m_{i}, \boldsymbol{\hat{C}}_i)\right\|_{2}-1)^{2}$ \;
        	Update $\theta^{Disc}$ with $\mathcal{L}_D$
        }
    $\hat{\mathcal{L}_G} = -\mathcal{F}_{Disc}(\boldsymbol{P}^o_{i}, \boldsymbol{P}^m_{i}, \boldsymbol{C}^g_i)$ \;
    $\mathcal{L}_G = \mathcal{L}_C(\boldsymbol{C}^g_i, \boldsymbol{C}_i^r) + \beta\hat{\mathcal{L}_G}$ \;
    Update $\theta^{LSG}$ with $\mathcal{L}_G$
 }
 \caption{Discriminative modeling algorithm}
 \label{ag:meta learning}
\end{algorithm}
Overall, the LSG model aims to summarize input video into high-level visual words to generate informative captions, and the discriminative modeling enhances the generated captions to be more semantically relevant.

\begin{table*}
\centering
\caption{Comparison between the proposed D-LSG and the state-of-the-art methods on MSVD and MSR-VTT datasets. B@4, M, R and C denote BLUE-4, METEOR, ROUGE-L and CIDEr, respectively.}
\begin{tabular}{lcccrrrrrrrr}
\toprule

\multicolumn{1}{l}{Method} &
\multicolumn{3}{c}{Features}  &
\multicolumn{4}{c}{MSVD}    &
\multicolumn{4}{c}{MSR-VTT}    \\ 
\cmidrule(lrr){2-4}
\cmidrule(lrrr){5-8}
\cmidrule(lrrr){9-12}

&
\multicolumn{1}{c}{Appearance}   &
\multicolumn{1}{c}{Motion}   &
\multicolumn{1}{c}{Region}   &
\multicolumn{1}{c}{B@4} &
\multicolumn{1}{c}{M}   &
\multicolumn{1}{c}{R}   &
\multicolumn{1}{c}{C}   & 
\multicolumn{1}{c}{B@4} &
\multicolumn{1}{c}{M}   &
\multicolumn{1}{c}{R}   &
\multicolumn{1}{c}{C}   \\ 

\midrule

PickNet \cite{chen2018less}   &	\checkmark& $\times$ &$\times$ & 52.3   & 33.3   & 69.6 & 76.5 & 41.3   & 27.7   & 59.8  & 44.1     \\
MARN \cite{pei2019memory}     &	\checkmark& 	\checkmark&$\times$ & 48.6   & 35.1   & 71.9 & 92.2 & 40.4   & 28.1   & 60.7 & 47.1     \\
OA-BTG \cite{zhang2019object} &	\checkmark&$\times$ &	\checkmark & 56.9   & 36.2   &  & 90.0  & 41.4   & 28.2   & - & 46.9\\
RMN \cite{tan2020learning}   &	\checkmark& 	\checkmark& 	\checkmark & 54.6   & 36.5   & 73.4 & 94.4 & 42.5   & 28.4   & 61.6 & 49.6\\
STG \cite{pan2020spatio}      &	\checkmark& 	\checkmark&	\checkmark & 52.2   & 36.9   & 73.9 & 93.0 & 40.5   & 28.3   & 60.9 & 47.1 \\
ORG-TRL \cite{zhang2020object} &	\checkmark& 	\checkmark& 	\checkmark& 54.3   & 36.4   & 73.9 & 95.2 & 43.6   & 28.8   & 62.1 & 50.9\\
C-R Reasoning \cite{hou2020joint} &$\times$&$\times$&\checkmark& 57.0   & 36.8   & - & 96.8 & - & - & - & -  \\
D-LSG                        &	\checkmark& 	\checkmark& 	\checkmark& \textbf{60.9}   & \textbf{37.6}   & \textbf{75.2} & \textbf{100.8} & \textbf{44.6} &  \textbf{28.8} & \textbf{62.3} & \textbf{51.2}\\  

\bottomrule
\end{tabular}

\label{tab:sota compare}
\end{table*}

\section{Experiments}
In this section, we present our experimental results on two public datasets: MSVD \cite{chen2011collecting} and MSR-VTT \cite{xu2016msr}. 
We compare our D-LSG with other state-of-the-art methods and an in-depth ablation study is provided to better understand our method.

\subsection{Experimental Setup}
\noindent\textbf{Datasets.} 
1) MSVD contains 1970 different YouTube short video clips with an average video length of 10.2s.
For each video, we used around 40 captions as only English was considered in all experiments.
Following \cite{tan2020learning}, we divided the dataset into three parts with 100 clips for validation, 1200 clips for training, and the remaining 670 clips for testing.
2) MSR-VTT is another dataset for open domain video captioning which consists of 10,000 video clips with an average video length of 14.8s and each of them is annotated with 20 English expressions. 
They are divided into 20 categories, such as music and movie. 
For fair comparisons, the standard splits are 6513 training videos, 497 validation videos and 2990 test videos.

\noindent\textbf{Evaluation Metrics. }
For a fair comparison, the quality of the generated captions in this study is evaluated by four evaluation metrics: BLEU-4 \cite{papineni2002bleu}, METEOR \cite{denkowski2014meteor}, CIDEr \cite{vedantam2015cider} and ROUGE-L \cite{lin2004rouge}. 
BLEU-4 measures the fraction of overlapping n-grams (here n = 4) between predicted sentences and reference sentences.
METEOR calculates the precision and recall between predicted sentences and references based on uni-gram, which extends exact word matching to various match levels.
CIDEr evaluates the consensus between a predicted sentence and reference sentences of the corresponding image or video based on the number of overlapping units such as n-gram.
ROUGE-L computes recall and precision scores of the longest common subsequences (LCS) between the generated and each reference sentence.
For all metrics, a higher value represents better performance of the generated captions.

\noindent\textbf{Data Preprocessing and Feature Extraction.}
We follow the process of \cite{tan2020learning} for corpus preprocessing and feature extraction. 
For corpus preprocessing, captions are first converted to lower case and punctuations are removed. 
Then, captions with more than 26 words are truncated and captions with less than 26 words are zero-padded. 
Besides, words that appear less than twice and five times are deleted in MSVD and MSR-VTT, respectively.
For feature extraction, 2D and 3D CNN feature extractors are InceptionResNetV2 (IRV2) \cite{szegedy2017inception} and I3D \cite{carreira2017quo}. 
Features from 26 frames are uniformly sampled in each video. 
Faster-RCNN \cite{ren2015faster} is adopted to extract the 36 region features for each frame out of the 26 sampled frames.

\noindent\textbf{Implementation Details.}
The Adam optimizer is applied with a learning rate $8 \times 10 ^{-4}$ for LSG model. For the discriminative model, we applied the Adam optimizer with ascent learning rates from $2 \times 10 ^{-4}$ to $8 \times 10 ^{-4}$. The size of hidden states for all LSTM models is 1024 and 1536 in MSVD and MSR-VTT datasets, respectively. The feature size for all graph operations is set to 1024 for both datasets. Layer normalization is applied on top of the LSTM layer and graph nodes to speed up convergence.
The word embedding size is set to 300 without the use of any pre-trained embedding such as glove.
Feature dimension in CNN for discriminative model word feature extraction is set to 512.
The training batch size is set to 128 for both datasets. Beam search is applied during inference with size 5.

\begin{table*}
\centering
\caption{Ablation Study of the proposed D-LSG on MSVD and MSR-VTT datasets. B@4, M, R and C denote BLUE-4, METEOR, ROUGE-L and CIDEr, respectively.
CGO only denotes the model only applies Conditional Graph Operation. LPA only indicates the model only applies Latent semantic Aggregation.}
\begin{tabular}{lcccrrrrrrrr}
\toprule

\multicolumn{1}{l}{Method} &
\multicolumn{3}{c}{Component}  &
\multicolumn{4}{c}{MSVD}    &
\multicolumn{4}{c}{MSR-VTT}    \\ 
\cmidrule(lrr){2-4}
\cmidrule(lrrr){5-8}
\cmidrule(lrrr){9-12}

&
\multicolumn{1}{c}{CGO}   &
\multicolumn{1}{c}{LPA}   &
\multicolumn{1}{c}{D-modeling}   &
\multicolumn{1}{c}{B@4} &
\multicolumn{1}{c}{M}   &
\multicolumn{1}{c}{R}   &
\multicolumn{1}{c}{C}   & 
\multicolumn{1}{c}{B@4} &
\multicolumn{1}{c}{M}   &
\multicolumn{1}{c}{R}   &
\multicolumn{1}{c}{C}   \\ 

\midrule

baseline&$\times$&$\times$&$\times$& 53.9   & 35.4   & 72.8 & 92.7 & 42.1   & 27.5   & 61.1  & 48.2     \\
CGO only&\checkmark&$\times$&$\times$& 56.5   & 36.8   & 73.6 & 96.1 & 43.2   & 28.2   & 61.9  & 50.3     \\
LPA only&$\times$&\checkmark&$\times$& 55.8   & 36.1   & 73.2 & 95.4 & 42.8   & 28.1   & 61.7 & 50.1     \\
LSG&\checkmark&\checkmark&$\times$& 57.9   & 37.2   & 74.1  & 99.4  & 44.5   & 28.5   & 62.0 & 50.7\\
D-LSG&\checkmark&\checkmark&\checkmark & \textbf{60.9}   & \textbf{37.6}   & \textbf{75.2} & \textbf{100.8} & \textbf{44.6}   & \textbf{28.8}   & \textbf{62.3} & \textbf{51.2}\\

\bottomrule
\end{tabular}

\label{tab:ablation study}
\end{table*}

\subsection{Quantitative Evaluation}
We compare our proposed D-LSG model with the state-of-the-art models on the MSVD and MSR-VTT datasets to evaluate our model's performance, and the results are listed in Table \ref{tab:sota compare}. 
The results illustrate that our model achieves the best performance on the MSVD and MSR-VTT datasets for all evaluation metrics, which indicates the effectiveness of our proposed model for the video captioning task.  
The detailed analysis of results on the MSVD and MSR-VTT datasets is shown below.

\noindent\textbf{Comparison with encoder-decoder models.}
We first compare our model with traditional encoder-decoder based models, including PickNet \cite{chen2018less} and MARN \cite{pei2019memory}.
We can observe from Table \ref{tab:sota compare} that the performance gains significant improvement, which means object information plays an important role in video captioning task.

\noindent\textbf{Comparision with object-based models.}
We then compare with several recent studies that consider detailed object information, including OA-BTG \cite{zhang2019object} and RMN \cite{tan2020learning}.
We can observe that our D-LSG gains better performance in all metrics on both datasets, proving the effectiveness of utilizing graph-based models to learn object-level features.
OA-BTG builds a bi-directional temporal graph based on object features. However, the interaction between different objects is less encoded in the graph structure. RMN utilizes an attention mechanism for encoding the object-level features. However, the object interactions are only considered in the motion modeling module. We observe that our D-LSG model provides more obvious improvements on CIDEr than BLUE-4, which indicates that modeling interactions between different objects help generate rich semantic captions.

\noindent\textbf{Comparision with GNN-based models.}
Finally, we compare our model with the most recent approaches that adopt GNN based methods, including ORG-TRL \cite{zhang2020object}, S-T Graph \cite{pan2020spatio}, and C-R Reasoning \cite{hou2020joint}.
Our model outperforms the mentioned GNN based models and achieves excellent performance in BLUE-4 and CIDEr metrics.
The BLUE-4 metric focuses on the fluency and logic of the generated captions and the CIDEr metric mainly focuses on content-relevant words in videos.
The performance proves that D-LSG successfully captures high-level semantic concepts.
In contrast, while ORG-TRL employs GCN to model object interactions, ORG-TRK does not take frame-level information into account when conducting object-level graph convolution. Our proposed LSG has better performance, indicating that aggregating object-level information conditioned on frame-level features helps the GNN learn better object representations.
C-R Reasoning utilizes extra datasets to build semantic knowledge graphs. On the contrary, we propose to use discriminative modeling to enhance the semantic extraction via reconstructing visual words. This demonstrates that D-LSG can succeed in extracting semantic information without using auxiliary databases.


\begin{figure}[t]

\centering
    \begin{subfigure}[b]{0.23\textwidth}
        \centering
        \includegraphics[width=0.95\textwidth]{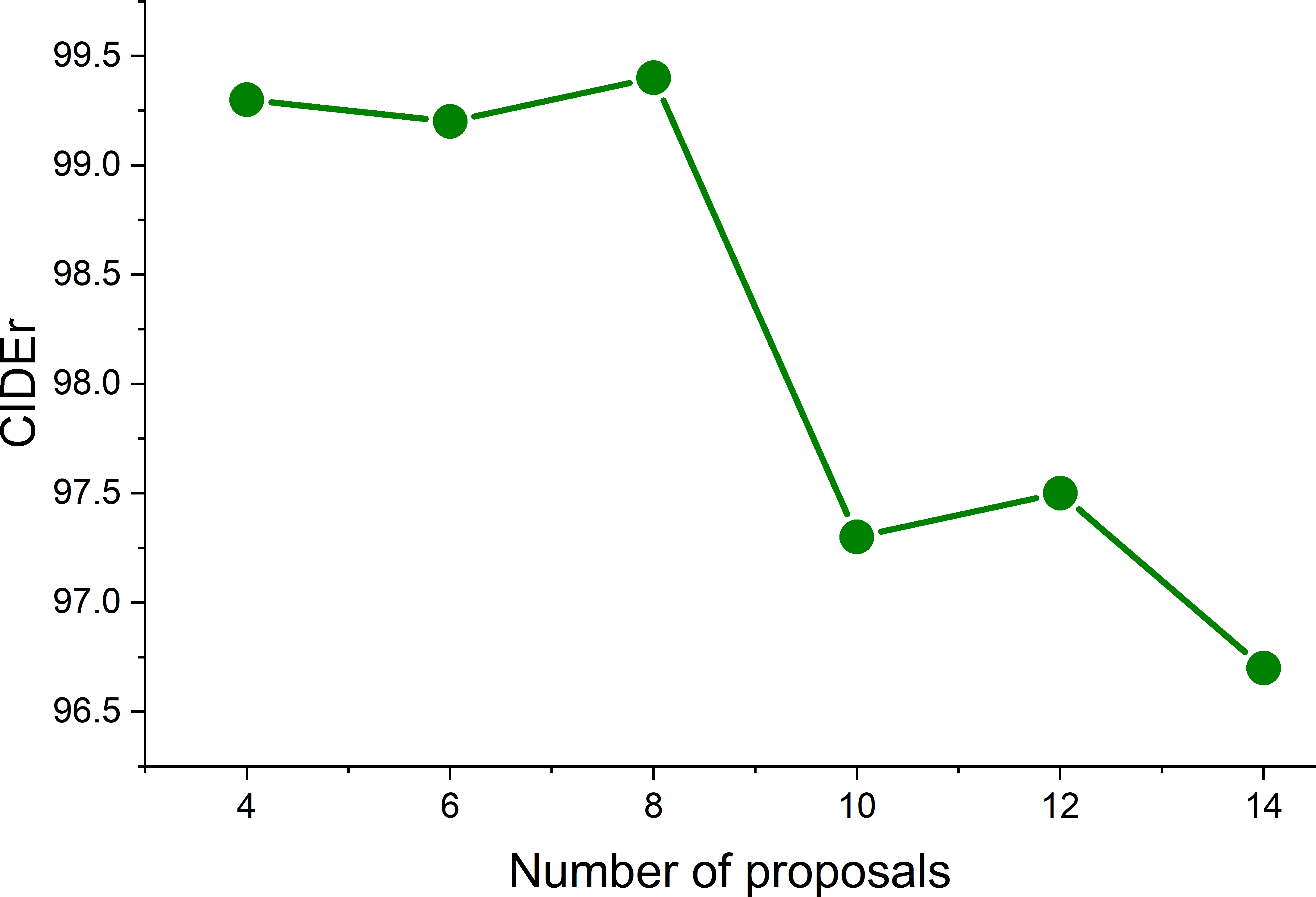}
        \caption{MSVD}
    \end{subfigure}%
    \begin{subfigure}[b]{0.23\textwidth}
        \centering
        \includegraphics[width=0.95\textwidth]{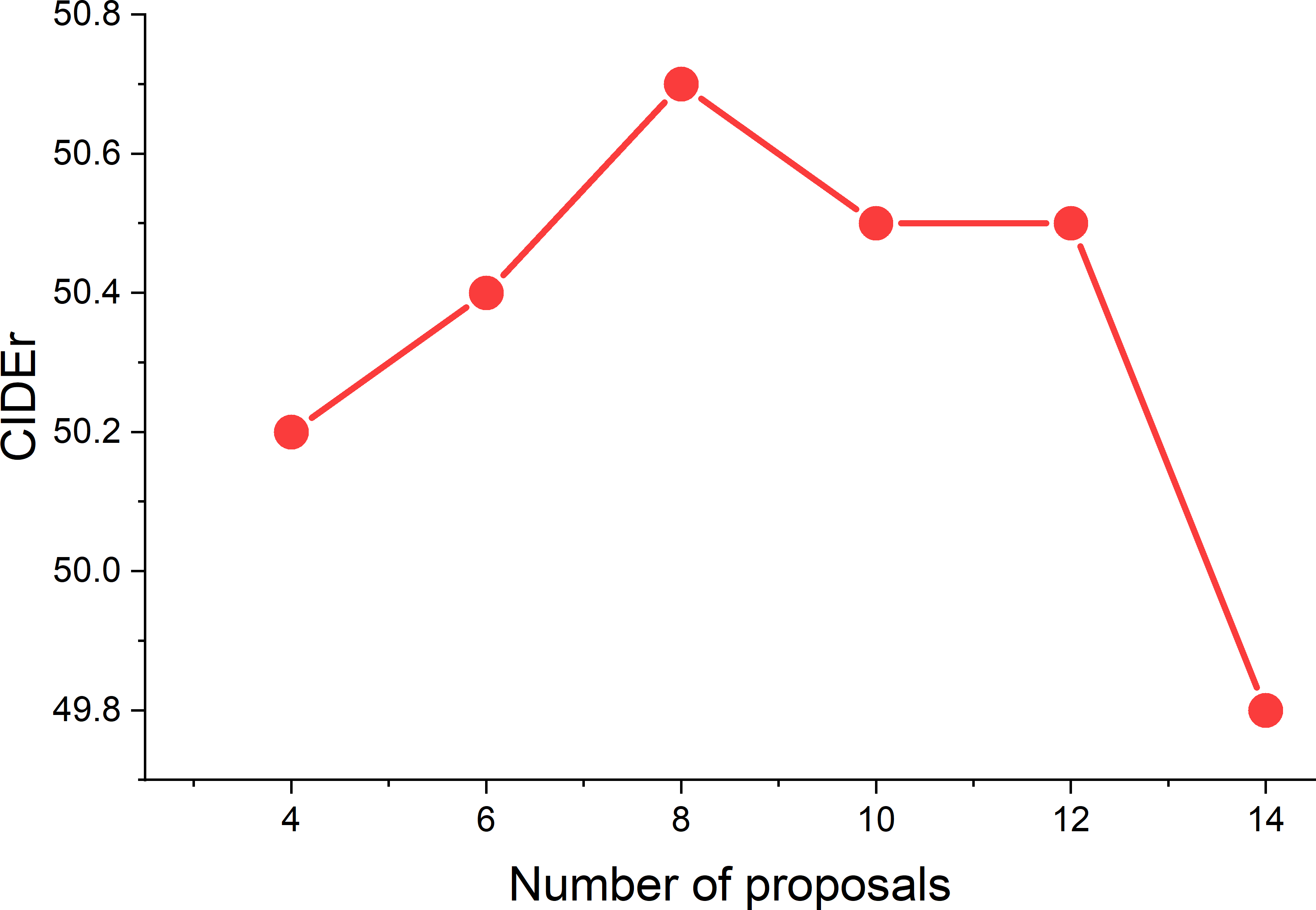}
        \caption{MSR-VTT}
    \end{subfigure}%
\caption{CIDEr of different visual words number in LSG on both MSVD and MSR-VTT datasets.}
\label{fig:psl number}
\end{figure}

\subsection{Ablation Study}
We then verify the effectiveness of the proposed D-LSG method through ablation studies on the MSVD and MSR-VTT dataset as shown in Table \ref{tab:ablation study}:
(1) baseline: the model inputs the concatenation of appearance feature and motion feature to the language decoder directly. (2) CGO only: the model only employs conditional graph operation. It aggregates the region features to frame-level appearance feature and motion feature, and feeds the enhanced object proposals directly to the language decoder without Latent Proposal Aggregation. (3) LPA only: the model summarizes the frame-level feature to visual words via Latent Proposal Aggregation without modeling object-level information. (4) LSG: the model includes the complete latent semantic graph. (5) D-LSD: the model combines LSG and the discriminative modeling part.
\begin{figure*}[ht!]
\begin{center}

\includegraphics[width=1.0\textwidth]{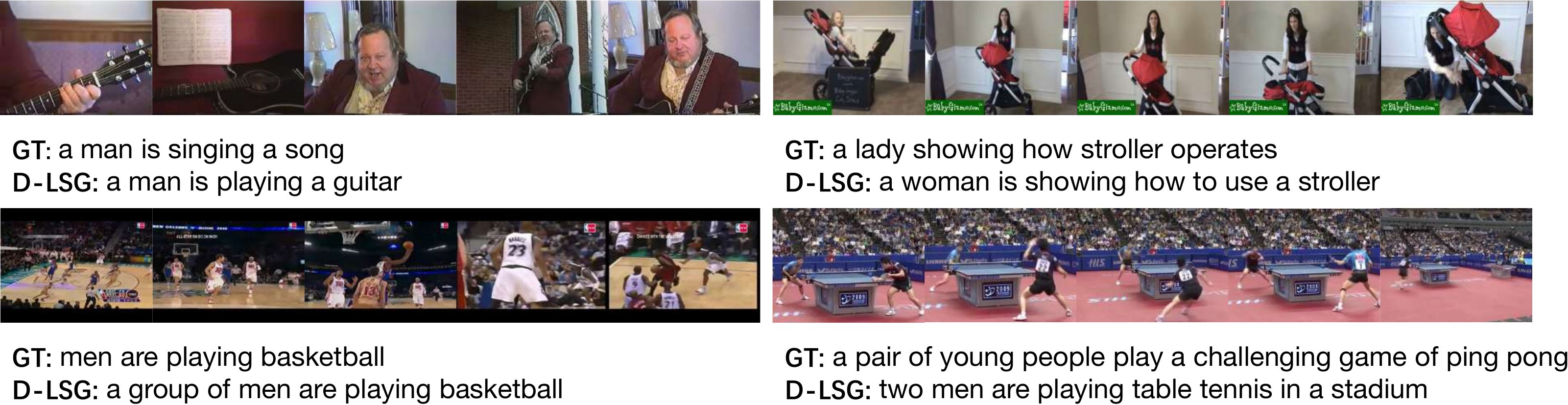}  
\end{center}
   \caption{Qualitative results of four videos from the MSVD and MSR-VTT datasets. The first line in each example is one of the ground truth captions and the second line is generated by our D-LSG method.}
\label{fig:caption_example}
\end{figure*}
\noindent\textbf{Effect of Graph.}
Comparing the results of CGO only and LSG, we observe a noticeable performance decrease on both datasets, which indicates the importance of summarizing frame-level features to latent concepts or visual words.
Comparing the results of LAP only and LSG, the performance also decreases. 
This is because LAP does not employ the conditional graph operation so that the visual words obtained in this model lack detailed object-level information. 
Since the performance drop of the LAP only model compared with the CGO only model is more pronounced, we can conclude that summarizing only frame-level information is not representative enough for semantic concepts, which also implies CGO's effectiveness modeling detailed object information. 


\noindent\textbf{Effect of latent proposal number.}
We also evaluated how the number of visual words affects the quality of the generated captions on the MSVD and MSR-VTT datasets. Figure \ref{fig:psl number} illustrates the performance on CIDEr using different numbers of visual words for both MSVD and MSR-VTT datasets.
As for the MSVD dataset, a small number of latent proposals provide better performance on CIDEr. However, when the number of proposals increases, the performance drops significantly. The intuition behind this is that the videos and captions in the MSVD dataset are short, so the data do not have enough semantic information to construct a large number of visual words, which results in performance drops. On the contrary, MSR-VTT with longer videos and captions suffers more performance drop when the number of proposals is small, which means that a small number of visual knowledge is not enough to represent the semantic concept of a given video. The above examples imply that The LSG model is able to summarize video content into semantic concepts with a proper number of visual words.

\noindent\textbf{Effect of discriminative modeling based on Graph.}
For the MSVD dataset, comparing LSG and D-LSG, we observe that METEOR and ROUGE-L have a slight improvement, while BLEU-4 and CIDEr show large improvements, especially for CIDEr. 
Though the advantage is less apparent on the MSR-VTT dataset, the increase of CIDEr is also noticeable when comparing the improvement on other evaluation metrics.
Since the mechanism of CIDEr is to punish words that are less informative of the video content, it may indicate that the discriminative structure can enrich the semantic concepts of the generated sentences, which means the model is capable of helping LSG capture and summarize key semantic concepts more effectively from input video features. 
The increase of BLEU-4 indicates that the discriminative model is capable of organizing enriched semantic concepts while keeping the grammatical structure of the generated captions.

\subsection{Qualitative Evaluation}

Figure \ref{fig:caption_example} shows examples of the generated captions on MSVD and MSR-VTT datasets. 
Compared to the ground truth captions, we can observe that the generated captions contain important objects (e.g.,``man'',``guitar'', ``stroller'') and motion (e.g., ``playing'', ``showing'') information even when the information is rare such as ``peeling'' and ``folding''. 
Also, the results indicate the visual words are capable of representing video content, and the model can capture frame-based background information rather than only focusing on detailed object information (e.g., ``stadium'' is detected in the example at bottom-right).





\section{Conclusion}
We have presented the first work to introduce graph neural networks and discriminative modeling to process spatio-temporal information for the video captioning task jointly. As for the Latent Semantic Graph, from the experiment results, we conclude that the Conditional Graph Operation effectively models detailed object-level interactions and relationships.
Besides, considering frame-level conditions is conducive to object-level interactive representation learning.
The Latent Proposal Aggregation component also succeeded in summarizing high-level visual knowledge from input video features. 
Also, the discriminative modeling enriched the generated captions' semantic information via visual knowledge reconstruction and discriminative training.
On two public datasets, our D-LSG model has outperformed the current state-of-the-art approaches, which verifies the effectiveness of our method.


\begin{acks}
Yu Guan is supported by Engineering and Physical Sciences Research Council (EPSRC) Project CRITiCaL: Combatting cRiminals In The CLoud (EP/M020576/1).
Yang Long is supported by Medical Research Council (MRC) Fellowship (MR/S003916/2).
\end{acks}

\newpage

\balance
\bibliographystyle{ACM-Reference-Format}
\bibliography{sample-base}

\end{document}